%% file: main.tex
\documentclass[runningheads]{llncs}

\usepackage[final,year=2024,ID=7447]{eccv}

\usepackage{eccvabbrv}

\usepackage{graphicx}
\usepackage{booktabs}
\usepackage{multirow}

\usepackage{amsmath}
\usepackage{algorithm}
\usepackage{algpseudocode}
\usepackage{subcaption}

\usepackage[accsupp]{axessibility}  

\newcommand{\mname}{ClusteringSDF\xspace}

\usepackage[pagebackref,breaklinks,colorlinks,citecolor=eccvblue]{hyperref}

\usepackage{orcidlink}

\begin{document}

\title{ClusteringSDF: Self-Organized Neural \\ Implicit Surfaces for 3D Decomposition}

\titlerunning{ClusteringSDF}

\author{Tianhao Wu\inst{1}$^\dagger$ \and
Chuanxia Zheng\inst{2} \and
Tat-Jen Cham\inst{1} \and 
Qianyi Wu\inst{3}}

\authorrunning{T.~Wu et al.}

\institute{Nanyang Technological University, $^\dagger$ S-Lab \and University of Oxford $~~~^3$ Monash University}

\maketitle
\vspace{-10pt}
\input{sec/0_abstract}
\input{sec/1_intro}
\input{sec/2_relative}
\input{sec/3_method}

\input{sec/4_exp}

\input{sec/5_limitation}
\input{sec/6_conclusion}

%
%
\bibliographystyle{splncs04}
\bibliography{egbib}

\clearpage

\appendix

\input{sec/sup}

\end{document}

%% file: sec/0_abstract.tex
\begin{abstract}
   3D decomposition/segmentation still remains a challenge as \emph{large-scale 3D annotated data is not readily available}. 
  Contemporary approaches typically leverage 2D machine-generated segments,
  integrating them for 3D consistency.
  While the majority of these methods are based on NeRFs, they face a potential weakness that the instance/semantic embedding features derive from independent MLPs, thus preventing the segmentation network from learning the geometric details of the objects directly through radiance and density.
  In this paper, we propose \mname, a novel approach to achieve both segmentation and reconstruction in 3D via the neural implicit surface representation, specifically Signal Distance Function (SDF), where the segmentation rendering is directly integrated with the volume rendering of neural implicit surfaces. 
  Although based on ObjectSDF++, \mname \emph{no longer requires the ground-truth segments for supervision} while maintaining the capability of reconstructing individual object surfaces, but purely with the noisy and inconsistent labels from pre-trained models.
  As the core of \mname, we introduce a high-efficient clustering mechanism for lifting the 2D labels to 3D and the experimental results on the challenging scenes from ScanNet and Replica datasets show that \mname can achieve competitive performance compared against the state-of-the-art with significantly reduced training time.
\keywords{3D segmentation \and neural implicit surface representation \and clustering}
\end{abstract}

%% file: sec/1_intro.tex
\section{Introduction}
\label{sec: intro}

\begin{figure}[h]
  \centering
  \includegraphics[width=\linewidth]{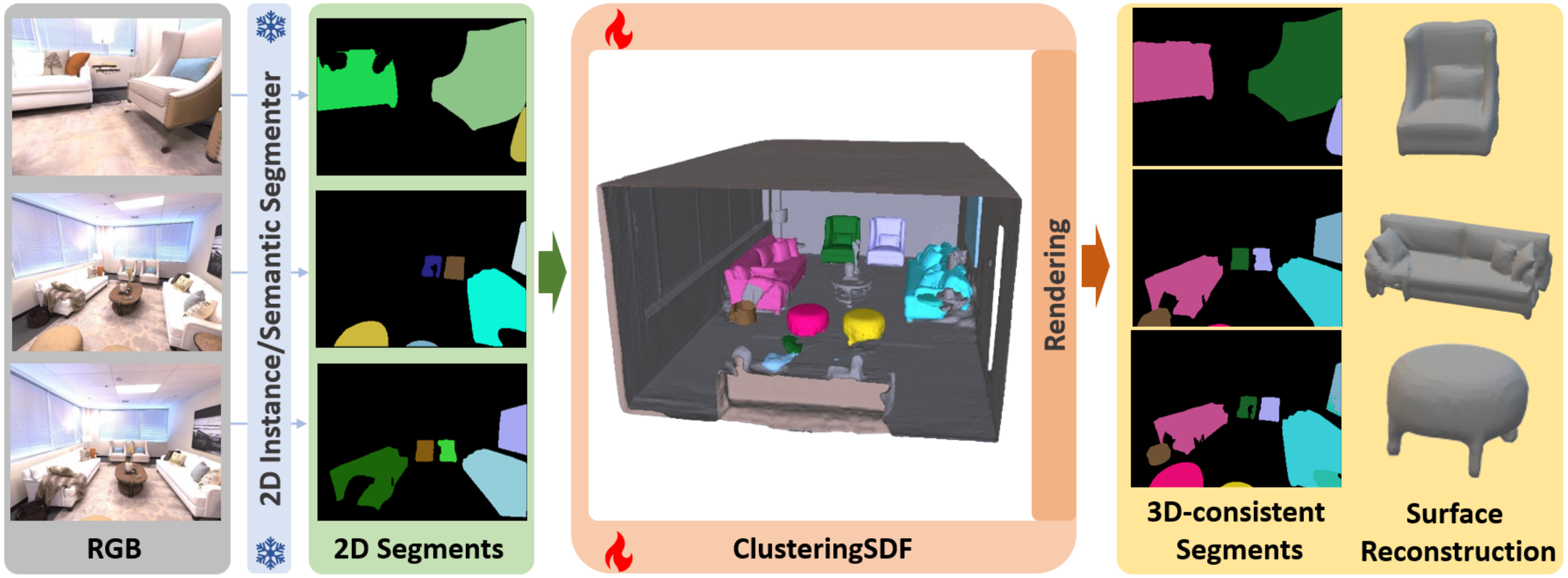}
  \caption{Taking RGB images from a scene and the corresponding 2D labels from a pre-trained segmentation model, our \mname is able to reconstruct the surface of the scene while fusing these inconsistent segments to be coherent and more accurate in 3D. Furthermore, it learns an object-compositional neural implicit representation and can reconstruct the surfaces of individual objects purely from these noisy labels.
  }
  \label{fig:overview}
\end{figure}

The vision community has rapidly improved 3D reconstruction and novel view synthesis (NVS) results with various neural implicit representations,
such as the Occupancy networks~\cite{mescheder2019occupancy},
DeepSDF~\cite{park2019deepsdf}, Scene Representation Network~\cite{sitzmann2019scene}, Neural Radiance Fields (NeRF)~\cite{mildenhall2021nerf}, and Light Field Network (LFN)~\cite{sitzmann2021light}.
Despite their remarkable performance, these systems ignore one critical fact: that the world is composed of individual 3D objects.
It is significant for multiple downstream applications, 
such as Augmented and Virtual Reality (AR \& VR) \cite{deng2022fov, li2022rt}, autonomous driving \cite{li2023read} and embodied AI \cite{liang2024av,byravan2023nerf2real},
to \emph{decompose and be able to interact with these 3D assets within a complex scene}.

At the same time, segmentation has undergone improvements~\cite{zhou2022detecting, cheng2022masked, kirillov2023segment, xu2023open} on 2D images in recent years.
However, recognizing and decomposing multiple 3D assets in a complex 3D scene is still a major challenge.
This is largely due to the complexity of \emph{processing and fusing 2D segmentation maps from multiple views into a consistent 3D representation},
especially when working with \emph{3D instances}. 
The accuracy of instance segmentation results, although potentially high, pertains exclusively to the current single view. 
However, labels assigned to each unique object in the scene \emph{do not remain consistent across views} \cite{ulku2022survey}.

In this paper, we aim to tackle the problem of \emph{object-centered} 3D reconstruction, 
wherein the goal is to \emph{simultaneously} enable 3D scene reconstruction, as well as \emph{3D assets decomposition or segmentation}.
Unlike previous approaches~\cite{zhi2021place,wu2022object,wu2023objectsdf++} that rely on ground-truth annotations,
our method leverages unaligned 2D segmentation results obtained from pre-trained models, fusing them to be consistent in the 3D representation.
Our \emph{key motivation} is that,
annotating precisely consistent 3D labels is challenging,
while the robust 2D segmentation models are widely available in the current vision zoo.
Although recent research~\cite{kundu2022panoptic, mascaro2021diffuser, kerr2023lerf, wang2022dm, siddiqui2023panoptic, bhalgat2023contrastive} have already focused on this area and have yielded good results in distilling 2D labels or features in 3D space,
they are mostly built upon a holistic NeRF~\cite{mildenhall2021nerf} for all instances within a scene.
A potential weakness lies in the disconnect between segmentation results and 3D reconstruction results, because the embedding features for semantic/instance rendering are derived from independent \texttt{MLPs} that do \emph{not} necessarily correlate with the underlying RGB or depth.
This separation makes it more difficult for the segmentation module to learn geometric details of the objects through radiance and density.

In contrast, neural implicit surface representations \cite{yariv2020multiview, oechsle2021unisurf, yariv2021volume, wang2021neus, wu2022object} that express 3D scenes as signed distance functions (SDFs) have the potential capability to resolve this segmentation challenge.
This is achieved by decomposing each instance as an independent \emph{object-level} surface,
and then further fusing all these 3D assets into a whole 3D scene (\cref{fig: overview}).
However, these SDF-based methods require precisely consistent labels for supervision, and currently, there is no proper solution to fuse the 2D information that is inconsistent across different views.

To address these aforementioned challenges, we propose a new neural implicit surface representation method, named \emph{\mname}, which is built upon ObjectSDF++ \cite{wu2023objectsdf++},
to lift the inconsistent 2D segmentation into consistent 3D assets.
While ObjectSDF++ \cite{wu2023objectsdf++} volume-renders object opacity masks and directly matches them to ground-truth labels, we replace this component entirely with our novel clustering method, so as to achieve segment assignments purely with the machine-generated 2D labels.
The crucial idea behind our \mname is to merge the predicted individual SDF channels from the implicit network, and treat them as SDF probability distributions or SDF features.
Instead of simply assigning each pixel to a specific predetermined SDF channel, the objective of \mname is to \emph{bring SDF distributions belonging to the same object closer, while concurrently separating those from different objects.}
The determination of whether pixels belong to the same object depends on their labels in the current view.
Compared against NeRF-based segmentation methods that predict radiance field densities, \mname renders neural implicit surfaces, where the gradients are concentrated around surface regions, leading to faster training and sharper segmented results.

In summary, our main contributions are as follows:
\begin{itemize}
    \item A novel \emph{self-organized} approach for fusing 2D machine-generated instance/\\semantic labels in 3D via neural implicit surface representation while simultaneously enabling holistic 3D reconstruction;
    \item A novel clustering mechanism for aligning these multi-view inconsistent labels, resulting in coherent 3D segment representations.
    Instead of computing similarity between each pair of pixels as contrastive loss, our method achieves high-efficient clustering by constraining the clustering centers within a simplex through normalization;
    \item Without ground-truth labels as supervision, our \mname maintains the object-compositional representations of the scene and its objects.
\end{itemize}

%% file: sec/2_relative.tex
\section{Related Work}
\label{sec: related}

\subsection{NeRFs for Semantic/Panoptic 3D Scene Modeling}
As one of the major approaches to achieve high-quality 3D reconstruction in recent years, the NeRF \cite{mildenhall2021nerf} learning-by-volume-rendering paradigm has led to considerable progress in the task of 3D scene understanding \cite{zhi2021place, fu2022panoptic, kundu2022panoptic, wang2022dm, kobayashi2022decomposing, tschernezki2022neural, kerr2023lerf}. 
Previous approaches have widely explored semantic 3D modeling and scene decomposition, from unsupervised foreground-background scene decomposition \cite{xie2021fig}, to leveraging weak signals like text or object motion \cite{mirzaei2022laterf, tschernezki2021neuraldiff, fan2022nerf}, and further to enhancing NeRF models with semantic labeling capabilities using annotations from 2D datasets \cite{sharma2023neural}.
SemanticNeRF \cite{zhi2021place} augmented the original NeRF to enable scene-specific semantic representation, by appending a segmentation renderer before injecting viewing directions into the MLP. 

Later works started focusing on the challenge of panoptic segmentation in 3D. 
DM-NeRF \cite{wang2022dm} learns unique codes of each identical object in 3D space from only 2D supervision, while Instance-NeRF \cite{liu2023instance}, by pre-training a NeRF and extracting its radiance and density field, utilizes 3D supervision to match inconsistent 2D panoptic segmentation maps.
Similarly, Panoptic-NeRF \cite{fu2022panoptic} leverages coarse 3D bounding primitives for panoramic and semantic segmentation of open scenes.
To decompose the scene into dynamic ``things'' and background ``stuff'' given a sequence of images, Panoptic Neural Fields (PNF) \cite{kundu2022panoptic} presents each individual object in the scene with a unique MLP and a dynamic track.

More recently, Panoptic Lifting \cite{siddiqui2023panoptic} and Contrastive Lift \cite{bhalgat2023contrastive} focused on directly lifting the 2D semantic/instance segment information without pre-training a 3D model or requiring any 3D bounding boxes.
Despite the impressive performance of these works, there remain some issues that have not been resolved.
While NeRF-based methods excel in rendering photorealistic 3D scenes, the semantic/instance embedding is not integrated directly with the learning of radiance (color) and density across the scene as they are derived from independent MLPs. 
Such deficiency makes it difficult for the model to learn the relationship between segmentation and 3D reconstruction, usually taking more training iterations to achieve a nuanced understanding of the scene.
On the other hand, the 2D machine-generated labels are inconsistent, with some over-segmented or under-segmented cases in the same scene. 
When the geometry information of the segmentation provided by these labels is inaccurate, such independence prevents the MLPs for segmentation from capturing the full geometric details of the objects through radiance and density, leading to inaccurate results.

\subsection{Object-Compositional Neural Implicit Surface Representation} 
Several research works \cite{wu2023objectsdf++, wu2022object, yariv2020multiview, yariv2021volume, wang2021neus} have revealed the powerful surface reconstruction ability of neural implicit surface representations. 
By implicitly representing 3D scenes as SDFs, the gradients are concentrated around the surface regions, leading to a direct connection with the geometric structure. 
On the contrary, for methods like Contrastive Lift \cite{bhalgat2023contrastive} and Panoptic Lifting \cite{siddiqui2023panoptic}, semantic rendering does not directly affect RGB or depth rendering, implying weak geometric constraints to supervise the correctness of segmentation.

Previous ObjectSDF \cite{wu2022object} and ObjectSDF++ \cite{wu2023objectsdf++} introduce mechanisms for representing individual objects within complex indoor scenes.
Compared to NeRF-based methods, the segmentation result has a direct impact on the final geometry rendering, including RGB, depth, and normals.
An excessive segmented region would require the model to render unrelated geometry at the corresponding location in space to satisfy such structure, while an incomplete segmentation would result in parts being missed from the geometric rendering, both of which will be penalized by the supervision.
However, they necessitate the input of ground-truth instance labels for accurate surface reconstruction, making it difficult to apply them to in-the-wild scene segmentation.
In this paper, we propose a new framework, based on ObjectSDF++ \cite{wu2023objectsdf++}, that incorporates a clustering alignment technique designed to fuse the inconsistent 2D instance/semantic labels.

\subsection{Clustering Operators for Segmentation}
Due to the fact that the labels of instances are almost entirely different across different camera viewpoints, relying directly on the values of the labels for alignment is inappropriate. 
Previous research \cite{zhao2021contrastive, de2017semantic, fathi2017semantic, kong2018recurrent, novotny2018semi} have explored clustering mechanisms for consistent 3D segmentation. 
Contrastive Lift \cite{zhao2021contrastive} attempts to make multi-view labels compatible using a contrastive loss function, together with a concentration loss between pixel-level embedding features. 
However, the computation of pixel-level contrastive loss can be time-consuming. 
To address this issue, we designed a more efficient way for clustering in this paper while achieving competitive segmentation results against the state-of-the-art models.

%% file: sec/3_method.tex
\section{Methodology}
\label{sec: method}

\begin{figure}[tb]
  \centering
  \includegraphics[width=\linewidth]{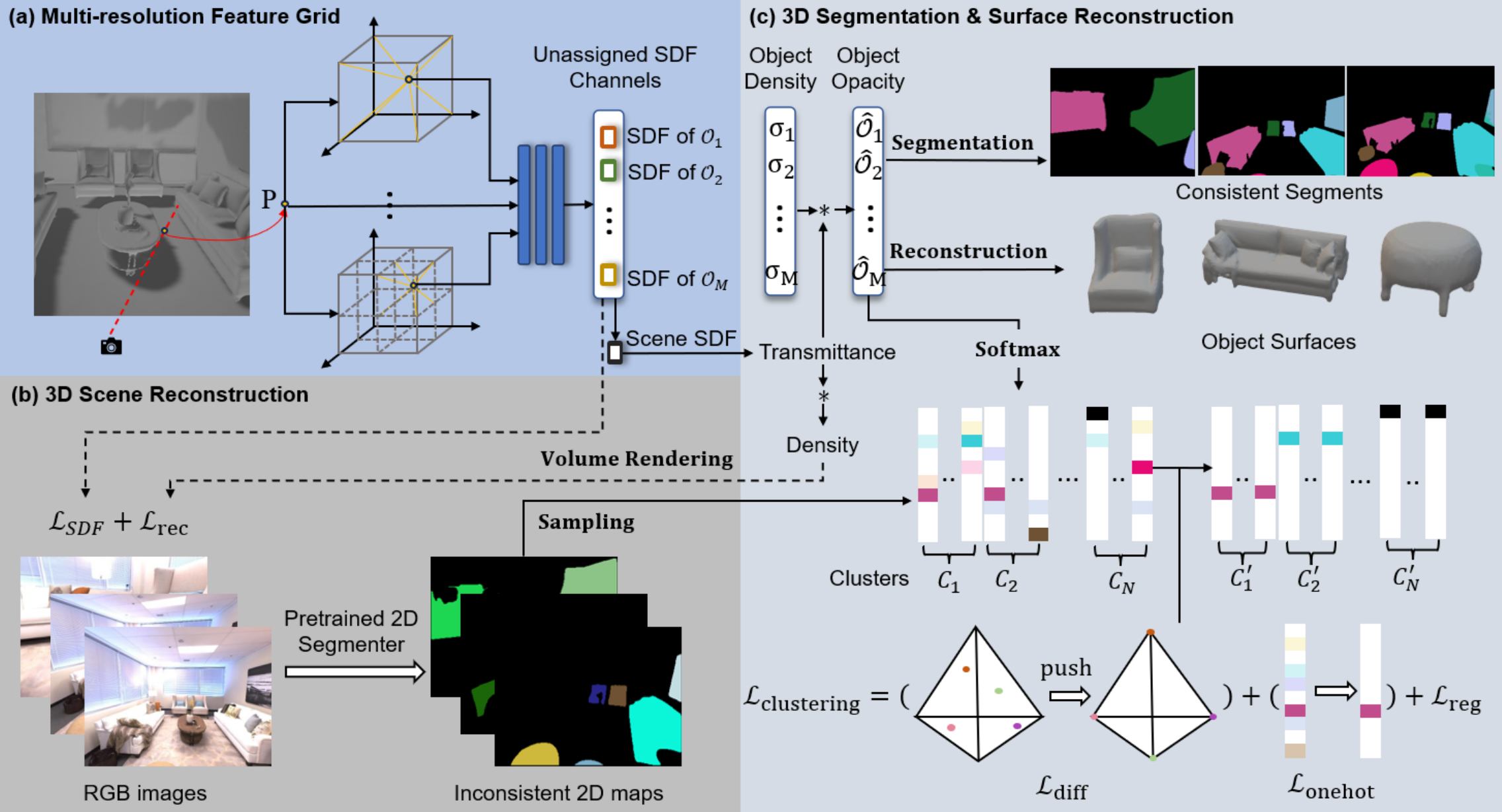}
  \caption{\textbf{Overview of \mname}. It is designed to fuse inconsistent 2D segments to 3D while reconstructing the object geometry by predicting their SDFs. To achieve this, we sample rays from single 2D segment maps and split them into $N$ groups corresponding to the $N$ distinct 2D labels $\{C_1,...,C_N\}$. The proposed $\mathcal{L}_{\text{diff}}$ then leverages normalized SDF distributions encompassing $c$ channels for individual objects (presented by different colors) and keeps the clustering centers to be apart, with $\mathcal{L}_{\text{onehot}}$ further encouraging the predicted clusters to be in the one-hot format.
  }
  \label{fig: overview}
\end{figure}

Given a set of $K$ posed RGB images $\mathcal{I}=\{I_1,...,I_K\}$ and the associated segmentation mask set $\mathcal{S}=\{S_1,...,S_K\}$ \emph{derived from a pre-trained 2D segmenter},
our main goal is to learn a model that not only reconstructs the whole 3D scene but also \emph{decomposes all 3D assets within it}.
Different from conventional methods that require precisely annotated semantic labels,
here we consider using \emph{labels from pre-trained 2D segmentation models}.
However, there are two main challenges if we directly use these machine-generated labels:
\textbf{i)} For any object in the scene, its corresponding segmented regions in different camera views are generally assigned different instance labels, even though it is the same object. 
\textbf{ii)} Furthermore, the estimated semantic labels can be noisy and inconsistent. 

To leverage these off-the-shelf labels from pre-trained 2D models, our \mname, illustrated in~\cref{fig: overview}, learns to align inconsistent 2D segmentation via a neural implicit surface representation framework. 
We first introduce an efficient clustering mechanism that overrides the need for precise object labels in \cref{sec:method_cluster}.
To refine the segmentation in 3D, we further design  one-hot and regularization terms to constrain the predicted value of each point in a single SDF channel, and a foreground-background loss to differentiate between foreground instances and background stuff in \cref{sec:method_fore}.
Moreover, in \cref{sec:method_semantic,sec:method_instance}, we introduce a semantic loss for matching the predetermined semantic labels, and a multi-view segmentation module to address the challenge of distinguishing between two or more objects that have never appeared together in any input image.

\subsection{Preliminaries}
\label{sec:method_pre}
\subsubsection{Neural Implicit Surface.}
In order to learn the SDF from multi-view images, a current approach~\cite{wang2021neus,yariv2021volume} is to replace the NeRF volume density output $\sigma(p)$ by a learnable transformation function of SDF value $d_\Omega(p)$, which is defined as the signal distance from point $\mathbf{p}$ to the boundary surface. 
Such an approach assumes that the objects have solid surfaces, and is supervised by reconstruction loss together with an Ekional regularization loss. 

\vspace{-8pt}\subsubsection{Occlusion-Aware Object Opacity Rendering.}
To further decompose the holistic scene representation into \emph{individual 3D assets}, the previous ObjectSDF~\cite{wu2022object} directly predicts an SDF for each object,
with minimum values across all such SDFs yielding the complete SDF for the scene.
More recently, ObjectSDF++~\cite{wu2023objectsdf++} has replaced the semantic-based cross-entropy loss by using an occlusion-aware object opacity,
which can be approximately expressed as:
\vspace{-4pt}
\begin{equation}
\hat{O}_{\mathcal{O}_i}(\mathbf{r})
=\int_{v_n}^{v_f}T_\Omega(v)\sigma_{\mathcal{O}_i}(\mathbf{r}(v))\,dv,
\quad i=1,...,M.
\label{eq: object opacity}
\end{equation}
Considering a ray $\mathbf{r}(v)$, the rendering opacity $\hat{O}$ for an object $\mathcal{O}_i$ can be expressed as an integral of its density $\sigma_{\mathcal{O}_i}$ and the scene transmittance $T_\Omega(v)$ along this ray from near bound ${v_n}$ to far bound ${v_f}$.
Here $M$ is the number of objects in the scene.
The surface representations for each object are then learned with supervision from ground truth instance segmentation masks:
\vspace{-4pt}
\begin{equation}
\mathcal{L}_O=\mathbb{E}_{\mathbf{r}\in\mathcal{R}}[\frac{1}{M}\sum_{i=1...M}\lVert\hat{O}_{\mathcal{O}_i}(\mathbf{r})-O_{\mathcal{O}_i}(\mathbf{r})\rVert]
\label{eq: object opacity loss}
\end{equation}
Here $\mathcal{R}$ is the set of rays sampled in a minibatch and $O_{\mathcal{O}_i}$ is the GT mask.

\subsection{Clustering Loss}
\label{sec:method_cluster}

Let $\hat{O}$ denote the set of predicted opacity values in a minibatch,
$y$ be the set of corresponding machine-generated 2D instance/semantic values in this batch and $L$ be the set of unique labels in $y$.
For each label $l\in{L}$, we define mask $m_l$ to identify the elements in $y$ belonging to the label $l$.
Given a set of rays $\mathcal{R}$ sampled \textbf{under the same view} in a minibatch,
which contains $N$ unique labels $l_1, l_2, ..., l_N$,
we split them into $N$ groups, denoted as $m_{l_1}, m_{l_2}, ..., m_{l_N}$. 
Instead of simply treating the predicted opacities as multiple individual SDF channels,
we adopt a perspective that, after \texttt{Softmax} normalization across different channels, 
\emph{they can instead be considered as probability distributions that assign each ray to different instance categories with different probabilities}.
Based on this idea, rather than optimising \cref{eq: object opacity loss}, we define a novel clustering loss:
\begin{equation}\mathcal{L}_{\text{clustering}}=\lambda_1\mathcal{L}_{\text{diff}}+\lambda_2\mathcal{L}_{\text{onehot}}+\lambda_3\mathcal{L}_{\text{reg}}
\label{eq: clustering loss}
\end{equation}
It consists of three main components: a differentiation loss, a one-hot loss and a regularization loss, which will be discussed in detail in this section.

The differentiation loss is formally defined as:
\vspace{-5pt}
\begin{equation}
\mathcal{L}_{\text{diff}}=-\frac{1}{N(N-1)}\sum_{i=1}^{N}\sum_{j=i+1}^{N}\lVert\mu(\hat{O}^s_{m_{l_i}})-\mu(\hat{O}^s_{m_{l_j}})\rVert_2,
\label{eq: diff loss}
\end{equation}
where $\hat{O}^s=\texttt{Softmax}(\hat{O})$, $\hat{O}_{m_{l*}}$ denotes the prediction set corresponding to set $m_{l*}$, 
and $\mu(\hat{O}^s_{m_{l_*}})$ is the clustering center of each category.
For the opacity predicted for each ray, which contains $c$ individual SDF channels, we consider the corresponding ($c$-1)-dim simplex in a high-dimension space.
The vectors, \ie the probabilities, can be seen as the points in this space that are constrained to lie within or on the surface of this simplex due to the \texttt{softmax} operation.
As illustrated in \cref{fig: overview}, \emph{the distance between the vertices is greater than the distance between any other pair of points inside a simplex}.
Hence our $\mathcal{L}_{\text{diff}}$,
which aims to maximize the L2 distances between each pair of clustering centers ($\mu(\hat{O}^s_{m_{l_i}})$ and $\mu(\hat{O}^s_{m_{l_j}})$),
will keep them apart while encouraging them to occupy different vertices of this simplex.
Therefore, it creates a gradient pushing the clustering centers to have only one channel to be 1 and the rest of the channels to be 0's.

For each pixel, our objective is to assign it to one of the SDF channels.
However, $\mathcal{L}_{\text{diff}}$ only takes into account the average predicted probabilities of each cluster, rather than the predictions for individual rays.
Based on this observation, we design the following one-hot loss $\mathcal{L}_{\text{onehot}}$ for further encouraging the model to produce approximate one-hot predictions for each pixel:
\vspace{-5pt}
\begin{equation}
\mathcal{L}_{\text{onehot}}= \frac{1}{|\mathcal{P}_{fg}|}\sum_{p_i\in{\mathcal{P}_{fg}}}\lVert\hat{O}_{p_i}-\text{onehot}(\hat{O}^s_{p_i})\rVert,
\label{eq: entropy loss}
\end{equation}
where $\mathcal{P}_{fg}$ are the pixels recognized as foreground objects in 2D segments. The \texttt{onehot} operation sets the largest-valued  channel in $\hat{O}^s_{p_i}$ to 1, and other channels to 0's.
However, our experiments show that the $\mathcal{L}_{\text{onehot}}$ causes the model to converge too fast and fall into local optima, leading to some cluster centers that should be different being pushed to the same vertices.
To allow the model to differentiate between different clusters at a more steady rate, we added an additional $\mathcal{L}_{reg}$ that encourages higher entropy by minimizing the \emph{variance of softmax values} in each cluster.
\vspace{-6pt}
\begin{equation}
\mathcal{L}_{\text{reg}}=\frac{1}{N}\sum_{i=1}^N\text{Var}(\hat{O}^s_{m_{l_i}}),\quad l_i\neq0.
\label{eq: same loss}
\end{equation}
Despite the fact that the objectives of $\mathcal{L}_{\text{onehot}}$ and $\mathcal{L}_{\text{reg}}$ contradict each other, by adjusting the weights, we are able to slow down the entropy reduction of the model, providing $\mathcal{L}_{\text{diff}}$ with enough time to separate the clustering centers.

Compared with the contrastive loss in~\cite{bhalgat2023contrastive}, $\mathcal{L}_{\text{clustering}}$ does \emph{not} focus on each positive/negative pixel pair, thus reducing the complexity from $O(N*|\mathcal{R}|^2)$ to $O(N^2*|\mathcal{R}|)$, where the number of unique labels $N$ is considerably smaller than the number of rays $|\mathcal{R}|$. 
The \texttt{Softmax} operation here plays a crucial role in transforming the predicted opacity into a probability distribution over the SDF channels.
Besides magnifying differences between channels and preventing the exploding or vanishing gradients, it inherently favors an unimodal distribution, where one class is significantly more probable than others, encouraging the model to be more decisive.

\subsection{Foreground-background Loss}
\label{sec:method_fore}
For 2D segmentation results, some objects are partially visible in some views, in which case they may not be segmented, while background stuff may be incorrectly segmented as foreground objects.
To better distinguish between background stuff and foreground objects, we can regulate the value of channel \#0 of the predicted opacity, which is the default channel to represent the background.
When the value of channel \#0 approaches 1, it indicates a higher probability that the pixel belongs to the background;
conversely, a lower value suggests a greater likelihood of being foreground. We design the following loss function to aid the model in distinguishing between foreground and background more rapidly:
\vspace{-6pt}
\begin{align}
\mathcal{L}_{\text{fg-bg}} = \lambda_4\mathcal{L}_{\text{bg}} + \lambda_5\mathcal{L}_{\text{fg}}=\lambda_4\frac{1}{|\mathcal{P}_{bg}|}\sum_{p_i\in{\mathcal{P}_{bg}}}|\hat{O}_{p_i0}-1|+\lambda_5\frac{1}{|\mathcal{P}_{fg}|}\sum_{p_i\in{\mathcal{P}_{fg}}}|\hat{O}_{p_i0}| 
\label{eq: bg loss}
\end{align}
By adjusting the values of $\lambda_4$ and $\lambda_5$, we can control how well the model differentiates between foreground objects and background stuff.

\subsection{Semantic Segmentation}
\label{sec:method_semantic}
To handle semantic segmentation, $\mathcal{L}_{\text{clustering}}$ alone is not sufficient, as it only requires that pixels of the same category be assigned to the same label, not that they correctly match the predefined semantic labels.
Therefore, we use an additional semantic loss to match cluster centers with one-hot vectors of the semantic categories, as derived from a pre-trained 2D generator.
\vspace{-6pt}
\begin{equation}
\mathcal{L}_{\text{sem}}=\frac{1}{N}\sum_{i=1}^{N}\mu(\mathcal{C}_{m_{l_i}})\lVert\mu(\hat{O}^s_{m_{l_i}})-\text{onehot}(y_i)\rVert_2,
\quad
l_i\neq0
\label{eq: semantic loss}
\end{equation}
Here $\mathcal{C}$ is the confidence map obtained from test-time augmentation, like as done in \cite{siddiqui2023panoptic}. By combining $\mathcal{L}_{\text{sem}}$ with our $\mathcal{L}_{\text{clustering}}$, the predicted semantic labels are efficiently aligned with the predetermined labels.

\subsection{Instance Segmentation}
\label{sec:method_instance}
Existing methods, including Contrastive Lift \cite{bhalgat2023contrastive} and ObjectSDF++ \cite{wu2023objectsdf++}, calculate the loss within the same view for instance segmentation. 
However, this leads to a potential problem: if two objects never appear in the same input image, they might end up being assigned to the same label.
To address this issue, we modify the sampling strategy from one single camera view to multiple views.
While 2D instance segmentation cannot support this multi-view supervision as the objects are generally labelled inconsistently between views,
we instead utilize 2D semantic segmentation to alleviate this problem, 
which provides both coherency and disambiguation across views.

Specifically, to avoid excessive increases in computation, we randomly select a small number of views, from which we sample a limited number of rays within a minibatch, while maintaining the previous single-view sampling setting.
Then we introduce a cross-view loss function to encourage the model to assign different instance labels to objects belonging to distinct semantic classes:
\vspace{-6pt}
\begin{equation}
\mathcal{L}_{\text{cross\_view}}=\exp(-\sum_{v\in\mathcal{V}}\sum_{\substack{l_1\in{L_c}\\l_2\in{L_v}}}\lVert\mu(\hat{O}^s_{m_{l_1}})-\mu(\hat{O}^s_{m_{l_2}})\rVert_2),
\quad
l_1\neq{l_2}
\label{eq: cross camera loss}
\end{equation}
Here $\mathcal{V}$ is the set of random sampled extra views, while ${L_c}$ and ${L_v}$ are the sets of unique \emph{semantic} labels under the current view and the extra view respectively. 

\subsection{Model Training}
\label{sec:method_training}
Except for the 2D segmentation fusion part, we continue to use the loss function of ObjectSDF++ \cite{wu2023objectsdf++} for color/normal/depth map $\mathcal{L}_{\text{rec}}$ and SDF reconstruction $\mathcal{L}_{\text{SDF}}$.
Therefore, the overall training loss for semantic segmentation:
\vspace{-6pt}
\begin{equation}
\mathcal{L}_{\text{semantic}}=\mathcal{L}_{\text{rec}}+\mathcal{L}_{\text{SDF}}+\mathcal{L}_{\text{clustering}}+\mathcal{L}_{\text{sem}}+\mathcal{L}_{\text{fg-bg}}
\label{eq: overall semantic loss}
\end{equation}
and for instance segmentation:
\vspace{-6pt}
\begin{equation}
\mathcal{L}_{\text{instance}}=\mathcal{L}_{\text{rec}}+\mathcal{L}_{\text{SDF}}+\mathcal{L}_{\text{clustering}}+\mathcal{L}_{\text{cross\_view}}+\mathcal{L}_{\text{fg-bg}}
\label{eq: overall instance loss}
\end{equation}

\subsection{Label-comprehensive Sampling and Key-frame Selection}
The baseline ObjectSDF++ \cite{wu2023objectsdf++} randomly samples a number of rays from the same camera view within a minibatch. 
However, in our \mname, random sampling does not guarantee the representation of every instance label, and the absence of one or more labels in the sampling process may lead to their omission in the model's training phase. 
This oversight results in slower convergence rates or an inability to correctly distinguish all labels in the final segmentation result.
In order to make it easier for the model to capture different labels, we adjusted the sampling strategy to ensure that the 1024 rays contain at least one of each of the different label values for the current view.
Furthermore, frames with more unique label values will be recognized as keyframes during the data loading phase and will be given extra attention by the model, with higher training frequency.

%% file: sec/4_exp.tex
\section{Evaluation}
\label{sec: exp}

\subsection{Experiment settings}
\subsubsection{Baselines.}
We performed a thorough comparison of \mname to the state-of-the-art segmentation methods, 
including SemanticNeRF \cite{zhi2021place}$_\text{NeurIPS'21}$, Panoptic Neural Fields \cite{kundu2022panoptic}$_\text{CVPR'22}$, Mask2Former \cite{cheng2022masked}$_\text{CVPR'22}$, DM-NeRF \cite{wang2022dm}$_\text{ICLR'23}$, Panoptic Lifting \cite{siddiqui2023panoptic}$_\text{CVPR'23}$, and Contrastive Lift \cite{bhalgat2023contrastive}$_\text{NeurIPS'23}$.

\vspace{-16pt}\subsubsection{Dataset.}
We assessed all models on ScanNet \cite{dai2017scannet} and Replica \cite{straub2019replica} datasets, following prior Panoptic Lifting \cite{siddiqui2023panoptic} and Contrastive Lift \cite{bhalgat2023contrastive}. The segmentation labels for supervision were derived from Mask2Former \cite{cheng2022masked}, while for quantitative metrics we compared the predicted results with the ground truth labels.

\vspace{-16pt}\subsubsection{Metrics.}
Following prior methods~\cite{siddiqui2023panoptic, bhalgat2023contrastive},
we report the standard metrics, including scene-level Panoptic Quality ($\text{PQ}^{\text{scene}}$) \cite{siddiqui2023panoptic} and mean intersection over union (mIoU).
For the ablation study, in order to make comparisons at a more granular level, we add two more metrics -- $\text{RQ}^{\text{scene}}$ \cite{siddiqui2023panoptic}, and our self-designed edge accuracy (EA).
Based on the standard Panoptic Quality (PQ) and Recognition Quality (RQ) \cite{kirillov2019panoptic}, $\text{PQ}^{\text{scene}}$ measures the scene-level consistency of all instance/stuff IDs across multiple views, while $\text{RQ}^{\text{scene}}$ measures the scene-level ID recognition accuracy.
To reflect the geometric accuracy, EA calculates the average Dice value between the edge of predicted and GT labels, which are paired by matching the densest labels in the predicted result with each GT label.

\vspace{-16pt}\subsubsection{Implementation Details.}
Our \mname is small enough to be trained on a single 12 GB GTX 2080Ti GPU. 
In particular, we train the model for 200 epochs, around 120k iterations on all scenes.
For each scene we train the model from scratch with~\cref{eq: overall semantic loss} and~\cref{eq: overall instance loss} separately.

\subsection{Main Results}

\begin{figure}[tb!]
  \centering
  \includegraphics[width=\linewidth]{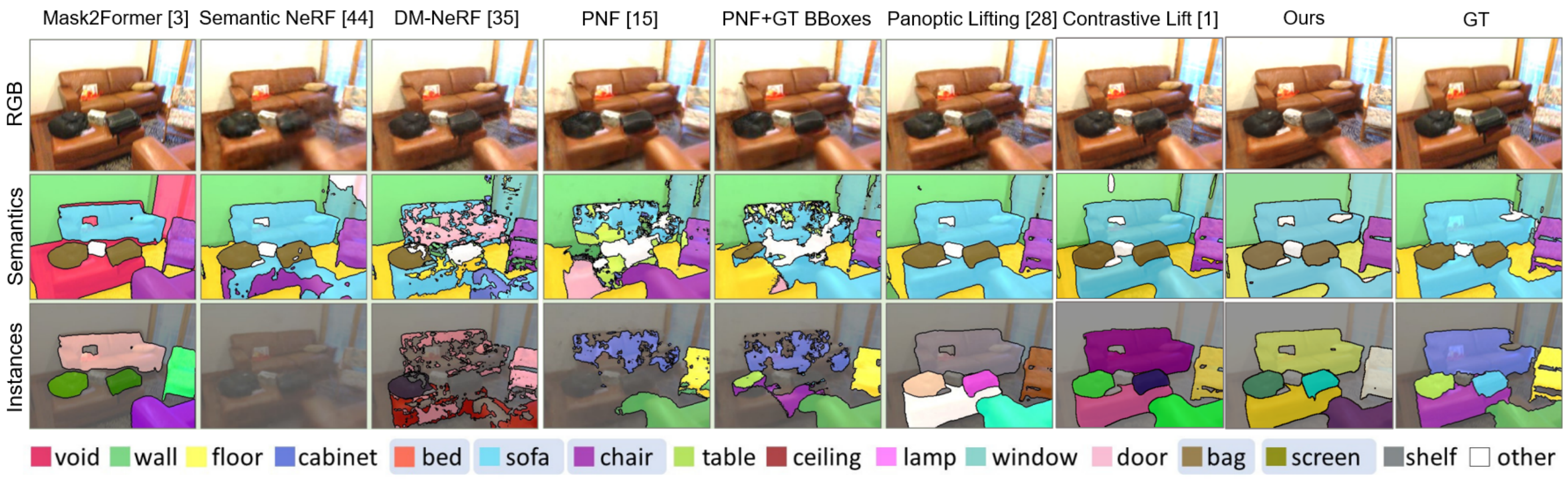}
  \vspace{-12pt}
  \caption{\textbf{Examples of semantic and instance segmentation on novel view.} Our \mname achieves very competitive results against previous 3D segmentation methods. Note that the instance and semantic results of our model are slightly different as we train two individual networks respectively. 
  Limited by the space, multi-view consistent segmentation results are provided in the supplementary materials.
  (Zoom in to see the details.)
  }
  \label{fig: qualitative comparison}
\end{figure}

\begin{figure}[tb!]
  \centering
  \includegraphics[width=\linewidth]{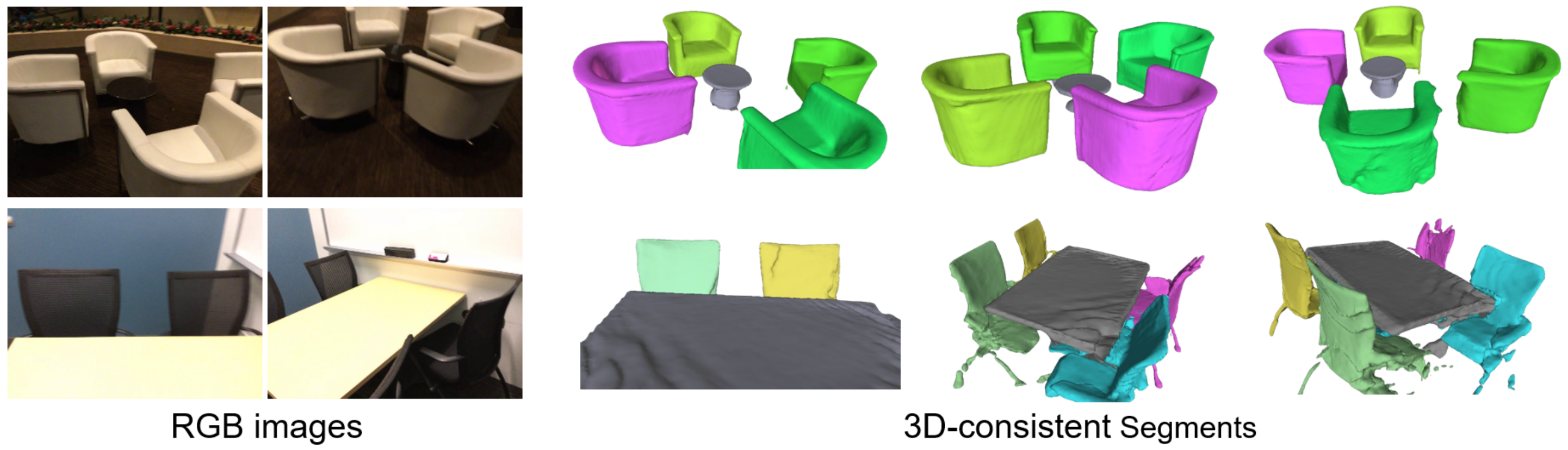}
  \vspace{-16pt}
  \caption{\textbf{Examples of our segmentation results in 3D.} Detail comparisons to existing methods on 3D consistency are provided in supplementary materials.
  }
  \label{fig: 3d consistency}
\end{figure}

\begin{table}[!tb]
  \caption{\textbf{Results on ScanNet and Replica datasets.} The performance of all prior work is sourced from \cite{siddiqui2023panoptic, bhalgat2023contrastive}. For each dataset, we report the $\text{PQ}^{\text{scene}}$ and mIoU metrics. 
  }
  \vspace{-4pt}
  \label{tab: quantitative}
  \centering
  \resizebox{0.6\linewidth}{!}{
  \begin{tabular}{@{}l ccc cc@{}}
    \toprule
     \multirow{2}{*}{\textbf{Method}} & \multicolumn{2}{c}{\textbf{ScanNet} \cite{dai2017scannet}} && \multicolumn{2}{c}{\textbf{Replica} \cite{straub2019replica}} \\
     \cline{2-3}\cline{5-6}
     & $\text{PQ}^{\text{scene}}$ $\uparrow$ & mIoU $\uparrow$ && $\text{PQ}^{\text{scene}}$ $\uparrow$ & mIoU $\uparrow$ \\
     \midrule
     Mask2Former \cite{cheng2022masked} & - & 46.7 && - & 52.4 \\
     SemanticNeRF \cite{zhi2021place} & - & 59.2 && - & 58.5 \\
     DM-NeRF \cite{wang2022dm} & 41.7 & 49.5 && 44.1 & 56.0 \\
     PNF \cite{kundu2022panoptic}  & 48.3 & 53.9 && 41.1 & 51.5 \\
     PNF + GT BBoxes & 54.3 & 58.7 && 52.5 & 54.8 \\
     Panoptic Lifting \cite{siddiqui2023panoptic} & 58.9 & 65.2 &&  57.9 & 67.2 \\
     Contrastive Lift \cite{bhalgat2023contrastive} & \textbf{62.3} & 65.2 && 59.1 & 67.0 \\
     \midrule
     \mname (\textbf{Ours}) & 61.0 & \textbf{66.2} && \textbf{59.3} & \textbf{68.9} \\
     \bottomrule 
  \end{tabular}
  }
\end{table}

We compare \mname to the state-of-the-art methods in 3D segmentation on ScanNet \cite{dai2017scannet} and Replica \cite{straub2019replica} in \cref{tab: quantitative}.
Here, we report both the $\text{PQ}^{\text{scene}}$ and mIoU metrics.
As shown in the table, the proposed \mname outperforms the baseline models on most of the metrics, except for the $\text{PQ}^{\text{scene}}$ on ScanNet \cite{dai2017scannet} against the state-of-the-art Contrastive Lift \cite{bhalgat2023contrastive}.
We believe it is mainly due to the relatively low quality RGB input of ScanNet \cite{dai2017scannet}, making it harder for our model to reconstruct the surface of the scene.
Especially on mIoU, which reflects the quality of the predicted semantic maps, our model outperforms the state-of-the-art methods by $+1.0$ and $+1.9$ points respectively.
These results demonstrate the effectiveness of our designed method.

The qualitative comparisons are visualized in \cref{fig: qualitative comparison}. 
\mname achieves good results \emph{without manually labeled dense semantic annotations}.
It is worth noting that our \mname can segment tiny objects better than the baseline models (\eg the pillow on the sofa).
Furthermore, to demonstrate that our segmentation results are consistent in 3D, we render the corresponding results on the mesh, seen as \cref{fig: 3d consistency}.
These results provide qualitative evidence that our model is effective for 3D segmentation and decomposition.

\begin{table}[!tb]
    \caption{ \textbf{Ablation study on Replica dataset.} To test the validity of each component of \mname, we remove each loss and test the performance. It should be noted that we use the same semantic segment results to highlight the difference in instance segmentation, which is more important for individual object representation.
    }
    \vspace{-4pt}
    \label{tab: ablation}
    \centering
    \resizebox{0.7\linewidth}{!}{
    \begin{tabular}{@{}l cc cc cc c@{}}
    \toprule
    Component & Supervision && $\text{PQ}^{\text{scene}}$ $\uparrow$ && $\text{RQ}^{\text{scene}}$ $\uparrow$ && EA $\uparrow$\\
    \toprule
    ObjectSDF++  & ground truth && 64.2 && 88.3 && 28.9 \\
    ObjectSDF++ w/ $\mathcal{L}_{\text{clustering}}$ & ground truth && \textbf{69.1} && \textbf{95.4} && \textbf{58.9} \\
    \midrule
    ObjectSDF++ & M2F \cite{cheng2022masked} && 43.8 && 55.3 && 12.9\\
    \mname w/o $\mathcal{L}_{\text{onehot}}$ & M2F \cite{cheng2022masked} && 56.3 && 61.3 && 23.2\\
    \mname w/o $\mathcal{L}_{\text{reg}}$ & M2F \cite{cheng2022masked} && 52.6 && 57.8 && 15.7\\
    \mname w/o $\mathcal{L}_{\text{fg-bg}}$ & M2F \cite{cheng2022masked} && 57.7 && 65.8 && 21.3\\
    \mname & M2F \cite{cheng2022masked} && \textbf{59.2} && \textbf{69.1} && \textbf{24.8} \\
    \bottomrule
    \end{tabular}}
\end{table}

\subsection{Ablation study}
We ran a number of ablations on Replica \cite{straub2019replica} to analyze the effectiveness of each core component in \mname. 
The results are shown in~\cref{tab: ablation} and~\cref{fig: ablation}.
Note that for semantic segmentation,
$\mathcal{L}_{\text{sem}}$ (\cref{eq: semantic loss}) plays an irreplaceable role for matching the labels, without which the semantic segmentation cannot be proceeded.
Therefore, we mainly explore the effect of our other proposed loss functions by probing their performance on instance segmentation, which directly impact the subsequent object surface reconstruction. 

\vspace{-8pt}\subsubsection{Clustering Loss.}
\begin{figure}[tb]
  \centering
  \subfloat[ObjectSDF++ with/without clustering loss.]{\includegraphics[width=0.53\textwidth]{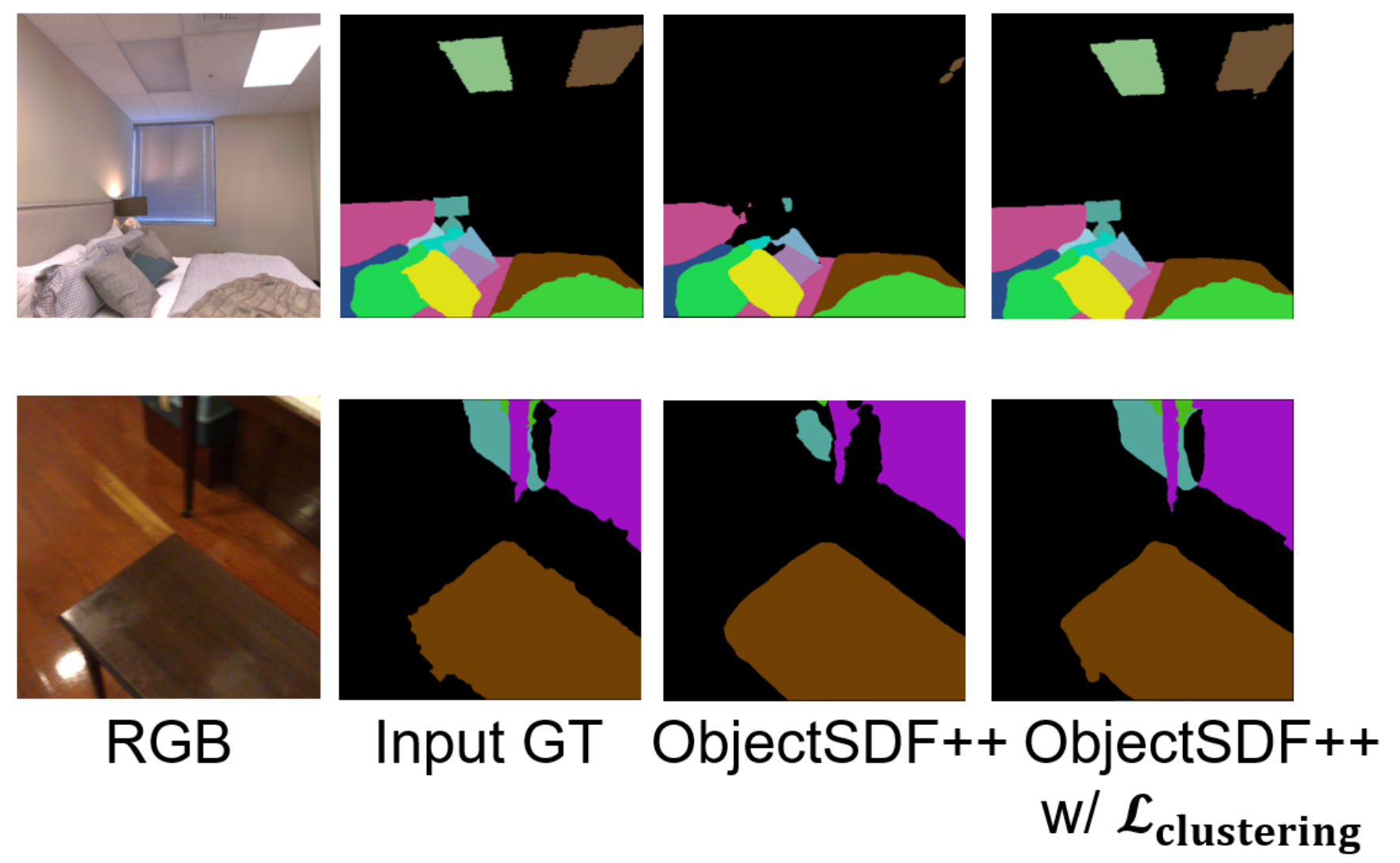}\label{fig: ablation sub1}}
  \hspace{0.1cm}
  \subfloat[Ablations over onehot, reg and fg-bg loss.]{\includegraphics[width=0.37\textwidth]{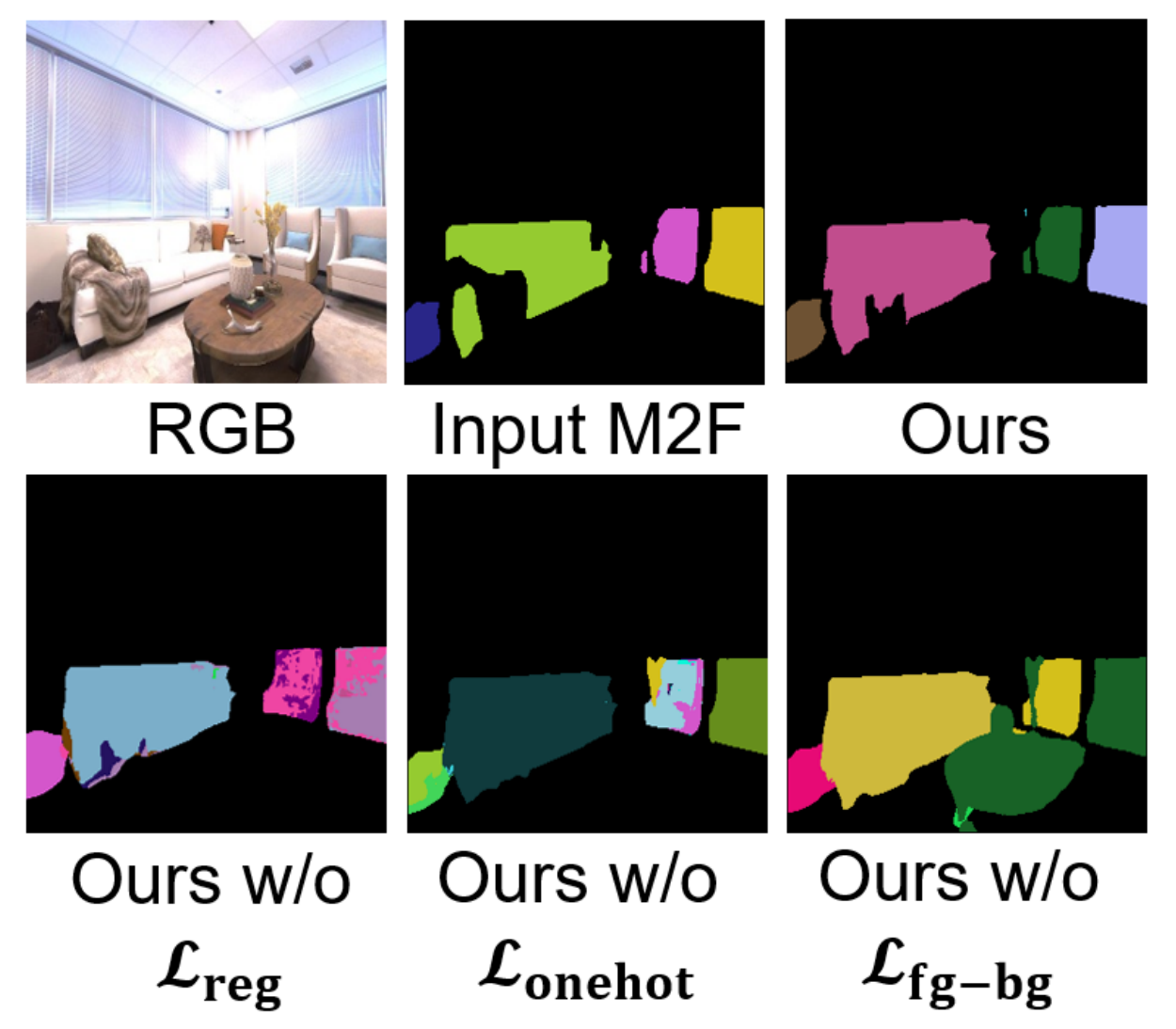}\label{fig: ablation sub2}} \\
  \subfloat[Instance differentiation with $\mathcal{L}_{\text{cross\_view}}$.]{\includegraphics[width=0.9\textwidth]{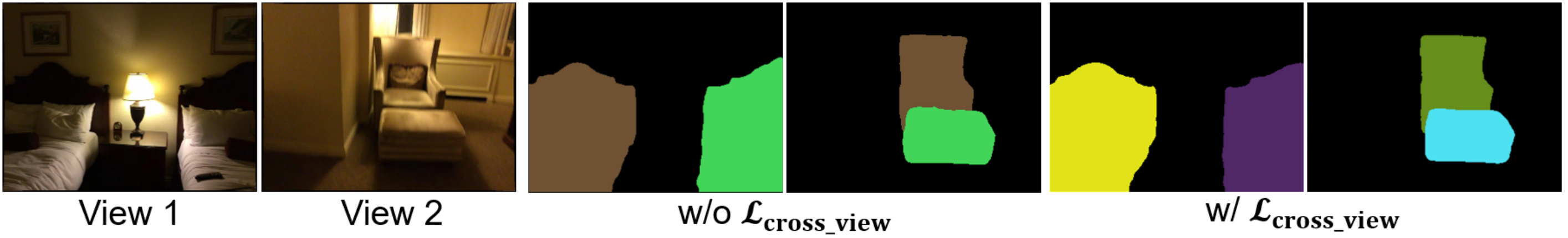}\label{fig: ablation sub3}}
  \vspace{-6pt}
  \caption{\textbf{Qualitative results for ablation study.}  
  }
  \label{fig: ablation}
\end{figure}

To investigate the efficiency of the primary component for instance segmentation -- $\mathcal{L}_{\text{clustering}}$ (\cref{eq: clustering loss}),
we integrate it into ObjectSDF++ \cite{wu2023objectsdf++} to assess whether it leads to further improvements when the GT instance labels are available.
The results are shown in the first two rows in~\cref{tab: quantitative},
where the overall performance demonstrates enhancement with the addition of $\mathcal{L}_{\text{clustering}}$.
Notably, the edge accuracy exhibits a substantial improvement of 30 points.
We attribute this to ObjectSDF++ being inferior at reconstructing planar objects, tending to integrate these objects into the background or into the surface of other objects.
Conversely, $\mathcal{L}_{\text{clustering}}$ substantially encourages the model to discriminate between different objects, thus enhancing the performance.
This improvement is further validated by the qualitative results depicted in~\cref{fig: ablation sub1}.

\vspace{-8pt}\subsubsection{One-hot Loss, Reg Loss and Fg-bg Loss.}
The loss term $\mathcal{L}_{\text{diff}}$ (\cref{eq: diff loss}) is also \emph{indispensable for segmentation}, omitting which the model will no longer work.
Therefore, we demonstrate the effect of our other designed loss terms $\mathcal{L}_{\text{onehot}}$, $\mathcal{L}_{\text{reg}}$ and $\mathcal{L}_{\text{fg-bg}}$ on different settings.
The quantitative results in \cref{tab: ablation} show that disabling each of these terms will lead to a substantial drop in the final performance.
We also provide rendering examples for more intuitive comparisons in~\cref{fig: ablation sub2}.
It shows that 
\textbf{i)} removing $\mathcal{L}_{\text{onehot}}$ significantly affects the smoothness, resulting in the same object being assigned with multiple labels; 
\textbf{ii)} the performance will be noticeably impaired without $\mathcal{L}_{\text{reg}}$ to constrain the speed of entropy reduction; 
\textbf{iii)} $\mathcal{L}_{\text{fg-bg}}$ encourages the model to distinguish between the foreground \emph{objects} and the background \emph{stuff}, without which they can be incorrectly segmented (\eg the table in the center should not be segmented).

\vspace{-8pt}\subsubsection{Cross-View Loss.}
Qualitative examples in \cref{fig: ablation sub3} shows that when different instances never appear together in one image, they can be potentially assigned with same labels (the left bed and the chair, the right bed and the footrest), while with our designed $\mathcal{L}_{\text{cross\_view}}$ they are correctly assigned with diverse labels. 

\subsection{Object Surface Reconstruction}
\begin{figure}[t]
  \centering
  \includegraphics[width=\linewidth]{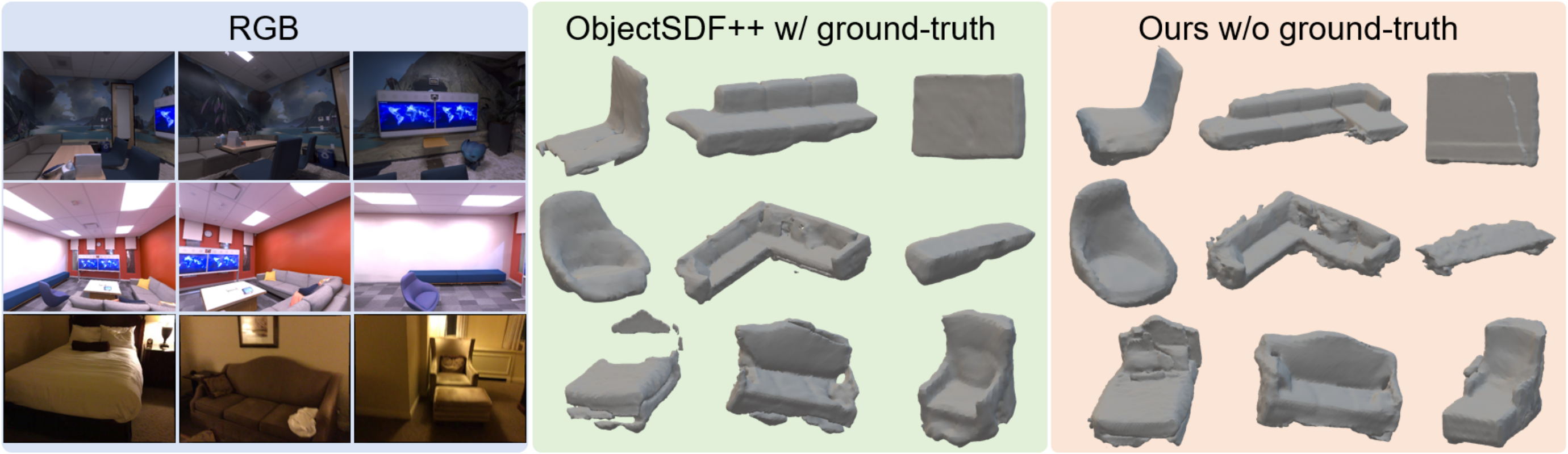}
  \vspace{-12pt}
  \caption{\textbf{Examples of object surface reconstruction.} Note that as ObjectSDF++ uses ground-truth labels for supervision, it imposes more detailed segmentation of some objects, resulting in some of the results appearing to be inferior to our approach.
  }
  \label{fig: surface reconstruction}
\end{figure}

As \mname is built upon object-compositional neural implicit surface representation,
while our $\mathcal{L}_{\text{onehot}}$ encourages the model to assign each pixel to distinct SDF channels,
it maintains the capability of reconstructing the surface of individual objects in the scene.
To validate this, in~\cref{fig: surface reconstruction}, we show examples of surface reconstruction on both 
ScanNet and Replica scenes. 
Compared against the reconstruction results of ObjectSDF++ \emph{trained with ground-truth labels}, our \mname can achieve nearly comparable reconstruction quality \emph{purely with 2D machine-generated labels}.

\subsection{Training time comparison}
Here we report the total iteration number and time needed to complete the training for the state-of-the-art \cite{bhalgat2023contrastive,siddiqui2023panoptic,wu2023objectsdf++} and \mname in \cref{tab: time comparison}. 
The results indicate that \mname can converge significantly more rapidly than the baseline methods while still achieving competitive or even better results.

\begin{table}[!tb]
  \caption{\textbf{Training time comparison.}}
  \vspace{-4pt}
  \label{tab: time comparison}
  \centering
  \resizebox{0.8\linewidth}{!}{
  \begin{tabular}{@{}lccc|clcc@{}}
  \toprule
  \textbf{Method} & iteration & training &&& \textbf{Method} & iteration & training\\
  \midrule
  Panoptic Lifting \cite{siddiqui2023panoptic} & 400K & >40h &&& ObjectSDF++ \cite{wu2023objectsdf++} & 200K & <12h\\
  Contrastive Lift \cite{bhalgat2023contrastive} & 400K & >30h &&& \mname (\textbf{Ours}) & 120K & <8h\\
  \bottomrule
  \end{tabular}}
\end{table}

%% file: sec/5_limitation.tex
\section{Limitations}
\label{sec: limitation}

While \mname shows good ability in fusing machine-generated 2D segmentation labels to 3D and reconstructing the object surfaces, there still remains some limitations. 
\mname is not able to handle semantic and instance segmentation simultaneously and requires separate training for now, thus leading to slight inconsistencies between the semantic and instance segmentation results.
To differentiate instances that had never appeared together in any input image, using semantic labels is still insufficient, as these instances may belong to the same semantic category.
In fact, the semantic labels here can easily be replaced with other supervisions as long as they are consistent across views, such as sparse manual labeling, which can be a practical direction for future development.

%% file: sec/6_conclusion.tex
\section{Conclusion}
\label{sec: conclusion}
We propose \mname, a novel approach for lifting 2D machine-generated segmentation labels to 3D via the object-compositional neural implicit surface representation while being able to simultaneously reconstruct the surfaces for the objects in the scene. 
Experimental results have shown that our model achieves state-of-the-art performance for rendering consistent 3D segmentation labels.
Through our designed high-efficient clustering mechanism, our \mname costs less than a quarter of the training time on each scene compared with the latest NeRF-based 3D segmentation methods. 
Furthermore, we show the capability of our approach in reconstructing surfaces for individual objects in the scene entirely from these inconsistent 2D labels, which we believe will provide inspiration for the subsequent research.

%% file: sec/sup.tex
\section{Appendix}

\setcounter{table}{0}
\setcounter{figure}{0}
\renewcommand{\thetable}{\Alph{table}}
\renewcommand{\thefigure}{\Alph{figure}}

In~\cref{sec:supp_impl}, we provide more implementation details.
In~\cref{sec:supp-proofs}, we provide a detail proofs of our clustering loss.
In~\cref{sec:supp-results}, we provide more visually qualitative results.


\subsection{Implement Details}
\label{sec:supp_impl}
\subsubsection{Training Parameters}
As mentioned in the main paper,
the attainment of $\mathcal{L}_{\text{clustering}}$ requires a meticulous adjustment of the weights among various components to show the model's entropy diminution.
Furthermore, within $\mathcal{L}_{\text{fg-bg}}$, it is imperative to control the model's proficiency in distinguishing between foreground objects and background stuff.
To this end, we set the hyperparameters $\lambda_1$, $\lambda_2$, $\lambda_3$ in clustering loss as 20, 0.5, 10000, and $\lambda_4$, $\lambda_5$ in foreground-background loss as 0.2, 0.1 during training.

\subsubsection{Key-frame Selection}
To enhance the stability of the training process,
it is imperative to encourage the model to focus more on the input images with more distinct labels, rather than dealing with single or even no segment labels.
To achieve this, when loading the entire training segmentation dataset, we will sort the label counts of each 2D segmentation map, \ie the number of different labels in each of the input segmentation maps.
Then we choose the top 10\% input maps with the most labels as key frames and those with no labels as empty frames.
During training, when empty frames are loaded, they will be randomly replaced with key frames.

\subsection{Proofs of Clustering Loss}
\label{sec:supp-proofs}
To further clarify how our novel-designed clustering loss works as we expect, we provide detailed proofs to illustrate the theory behind our design.

\subsubsection{Maximum Distance in a Probability Simplex}
\label{sec: maximum distance}
Considering $p,q\in{\mathbb{R}^N},~s.t.~\forall i\in{[1, N]},~p_i,q_i\geq{0}$, and $\sum_{i=1}^Np_i=1,~\sum_{i=1}^Nq_i=1$, the $\text{L}_2$ distance between $p$ and $q$ is:
\begin{equation}
    \lVert p-q \rVert_2^2=(p-q)^\top(p-q)=\lVert p \rVert_2^2+\lVert q \rVert_2^2-2p^\top{q},
\end{equation}

Since $\forall i\in{[1, N]},~p_i,q_i\geq{0}$, there are:
\begin{equation}
    (\sum_{i=1}^Np_i)^2\geq{\sum_{i=1}^Np_i^2}\Rightarrow\lVert p \rVert_2^2\leq{1},~(\sum_{i=1}^Nq_i)^2\geq{\sum_{i=1}^Nq_i^2}\Rightarrow\lVert q \rVert_2^2\leq{1}
\end{equation}
where the equality holds only if there is one and only one $p_i\in{p}$ and $q_j\in{q}$ is non-zero. As $\sum_{i=1}^Np_i=1,~\sum_{i=1}^Nq_i=1$, the non-zero $p_i$ and $q_j$ must be 1, \ie both $p$ and $q$ are in \textbf{one-hot} format. Furthermore, since $\forall_i{p_i\geq{0},q_i\geq{0}}$, we have $p^\top{q}\geq{0}$, with the equality holds only if $p$ and $q$ are \textbf{orthogonal}.

Therefore, $\lVert p-q \rVert_2^2 \leq 2$ and the bound is reached only when $p$ and $q$ are one-hot and orthogonal at the same time, \ie they occupy two different vertices of the simplex, which is the main objective of $\mathcal{L}_{\text{diff}}$.

\subsubsection{Minimum Variance in a Probability Simplex}
In order to slow down the entropy reduction and provide more time to separate the clusters, we designed $\mathcal{L}_{\text{reg}}$, which minimizes the variance of softmax values in each cluster. Here we demonstrate how it serves such an objective. Considering $p$ described same as in \cref{sec: maximum distance}:
\begin{equation}
\mu(p)=\frac{1}{N}\sum_{i=1}^N(p_i)=\frac{1}{N}
\end{equation}
\begin{equation}
\text{Var}(p)=\frac{1}{N}\sum_{i=1}^N(p_i-\mu(p))^2=\frac{1}{N}\lVert p-\frac{1}{N}\mathbf{1}\rVert_2^2
\end{equation}
$\text{Var}(p)$ is minimized when $p=\frac{1}{N}\mathbf{1}$, where $\frac{1}{N}\mathbf{1}$ is a uniform distribution of the same size as $p$ and all elements are $\frac{1}{N}$. Therefore, $\mathcal{L}_{\text{reg}}$ encourages an increase in the entropy of the predicted result $p$.

\subsection{Additional Results and Examples}
\label{sec:supp-results}

\subsubsection{Results of ObjectSDF++ with Machine-Generated Labels}

\begin{figure}[tb]
  \centering
  \includegraphics[width=\linewidth]{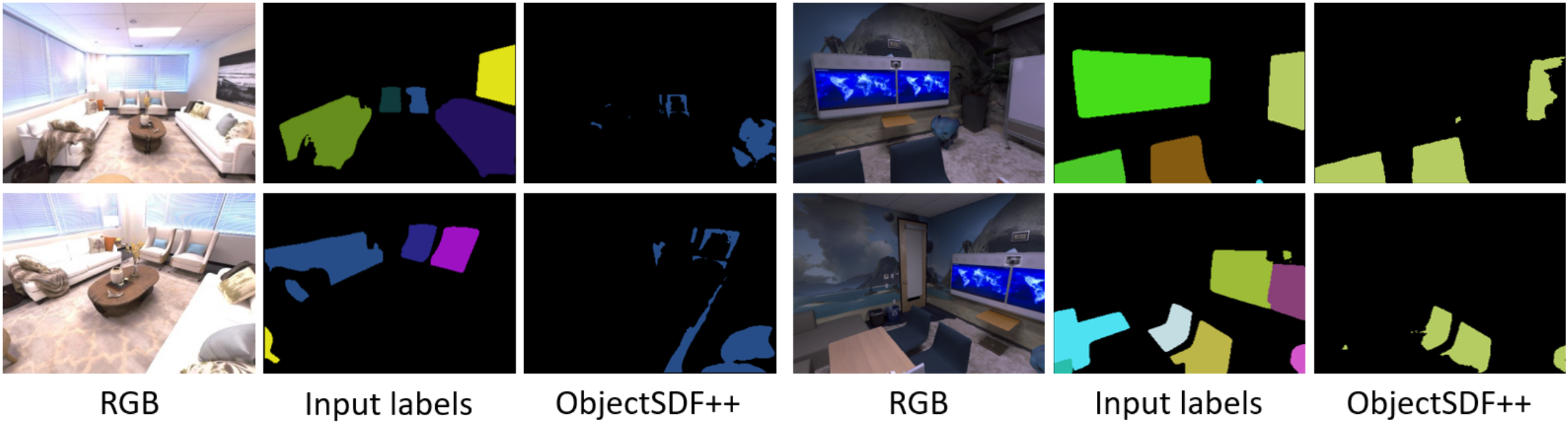}
  \caption{\textbf{Qualitative results of ObjectSDF++ with machine-generated instance labels}.
  }
  \label{fig: objsdf++}
\end{figure}

This is an extension of \cref{tab: ablation}, which reports only the quantitative results of training ObjectSDF++ \cite{wu2023objectsdf++} using the instance labels generated with Mask2Former \cite{cheng2022masked}.
Herein, we present the corresponding qualitative results in \cref{fig: objsdf++}.
As we can see, the original ObjectSDF++ cannot handle these view-inconsistent labels, resulting in many objects not being able to be segmented, or almost all objects being labeled the same way.

\subsubsection{Qualitative Results without Semantic-Matching Loss}
\begin{figure}[tb]
  \centering
  \includegraphics[width=0.7\linewidth]{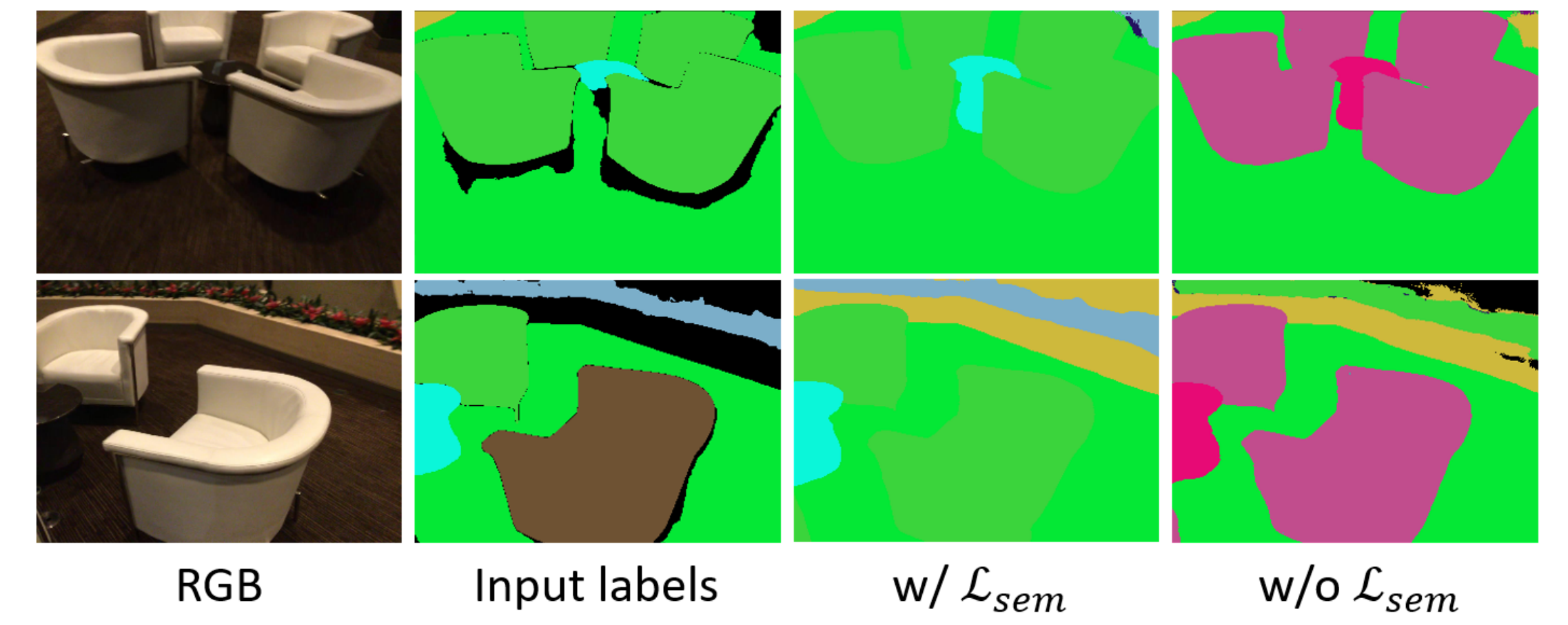}
  \caption{\textbf{Qualitative results without Semantic-Matching Loss}.
  }
  \label{fig: sem loss}
\end{figure}

In \cref{sec: exp}, we highlighted the indispensable role of $\mathcal{L}_{\text{sem}}$ in matching the labels, 
without which the semantic segmentation cannot proceed.
To demonstrate this, in \cref{fig: sem loss} we provide the results where the $\mathcal{L}_{\text{sem}}$ is omitted.
These results demonstrate that while the model retains the capacity to differentiate between objects of various categories, it \emph{fails to align the predicted labels with the predefined semantic labels accurately}.

\subsubsection{Qualitative Results without Differentiation Loss}
\begin{figure}[tb]
  \centering
  \includegraphics[width=\linewidth]{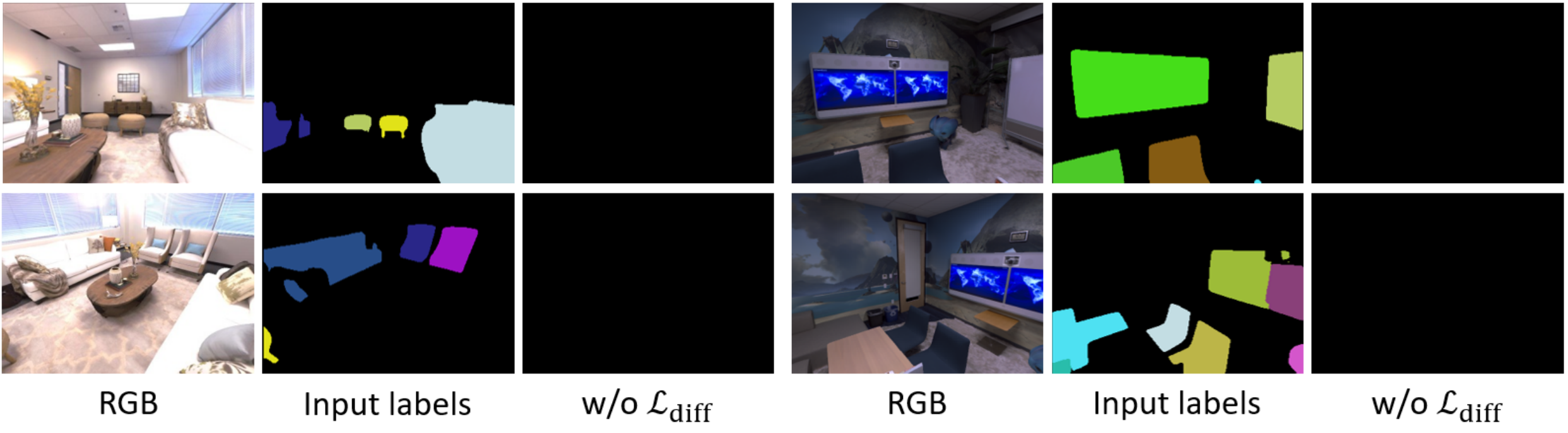}
  \caption{\textbf{Qualitative results of without $\mathcal{L}_{\text{diff}}$}.
  }
  \label{fig: diff loss}
\end{figure}

We also design another variant (w/o the loss term $\mathcal{L}_{\text{diff}}$) of the model,
because we also claimed it indispensable in \cref{sec: exp}.
Here, we provide the corresponding results in \cref{fig: diff loss} to prove this statement.
The qualitative results show that without $\mathcal{L}_{\text{diff}}$, the model will no longer work and cannot segment any objects.

\subsubsection{Comparison of Training Results at Early Stage}
\begin{figure}[t]
  \centering
  \includegraphics[width=\linewidth]{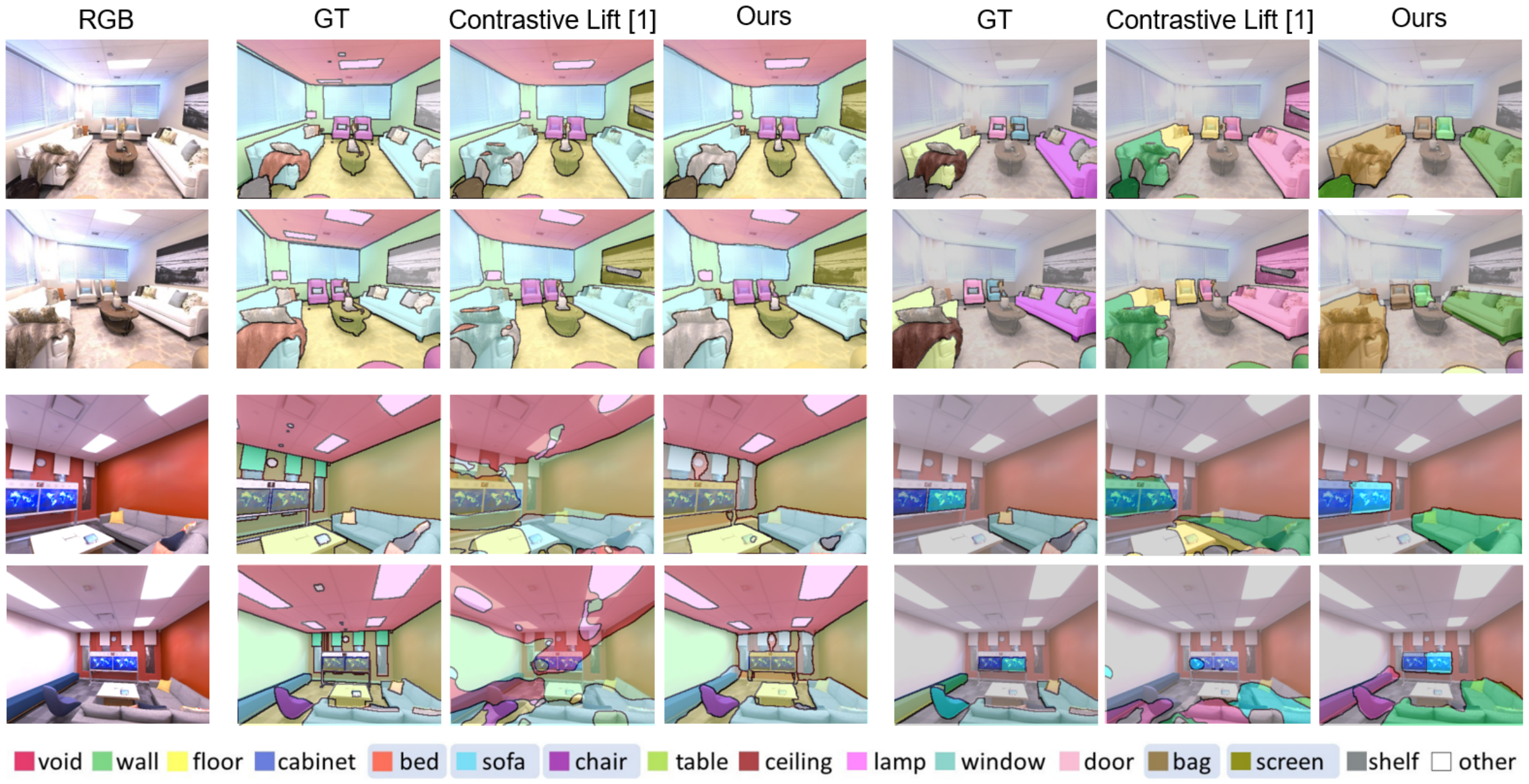}
  \caption{\textbf{Segmentation results at early training stage against Contrastive Lift \cite{bhalgat2023contrastive}.} Columns 2-4: semantic segmentation. Columns 5-7: instance segmentation. Please zoom in to see the details.
  }
  \label{fig: early training}
\end{figure}

To further demonstrate the efficiency of our method, we compare the segmentation results with the current state-of-the-art model, Contrastive Lift \cite{bhalgat2023contrastive} in the early stage of training. The results are shown as \cref{fig: early training}. Here we train our \mname and Contrastive Lift for less than 10\% of the complete training time, 3 hours for Contrastive Lift and 30 minutes for our model. Qualitative results show that our model can achieve sharp semantic/instance segmentation results at a very early stage. In contrast, for Contrastive Lift \cite{bhalgat2023contrastive}, 10\% of the training has apparently not yet enabled a relatively stable segmentation, especially for instance segmentation, which is significantly worse than our model.

\subsubsection{Multi-View Comparisons}
\begin{figure}[t]
  \centering
  \includegraphics[width=\linewidth]{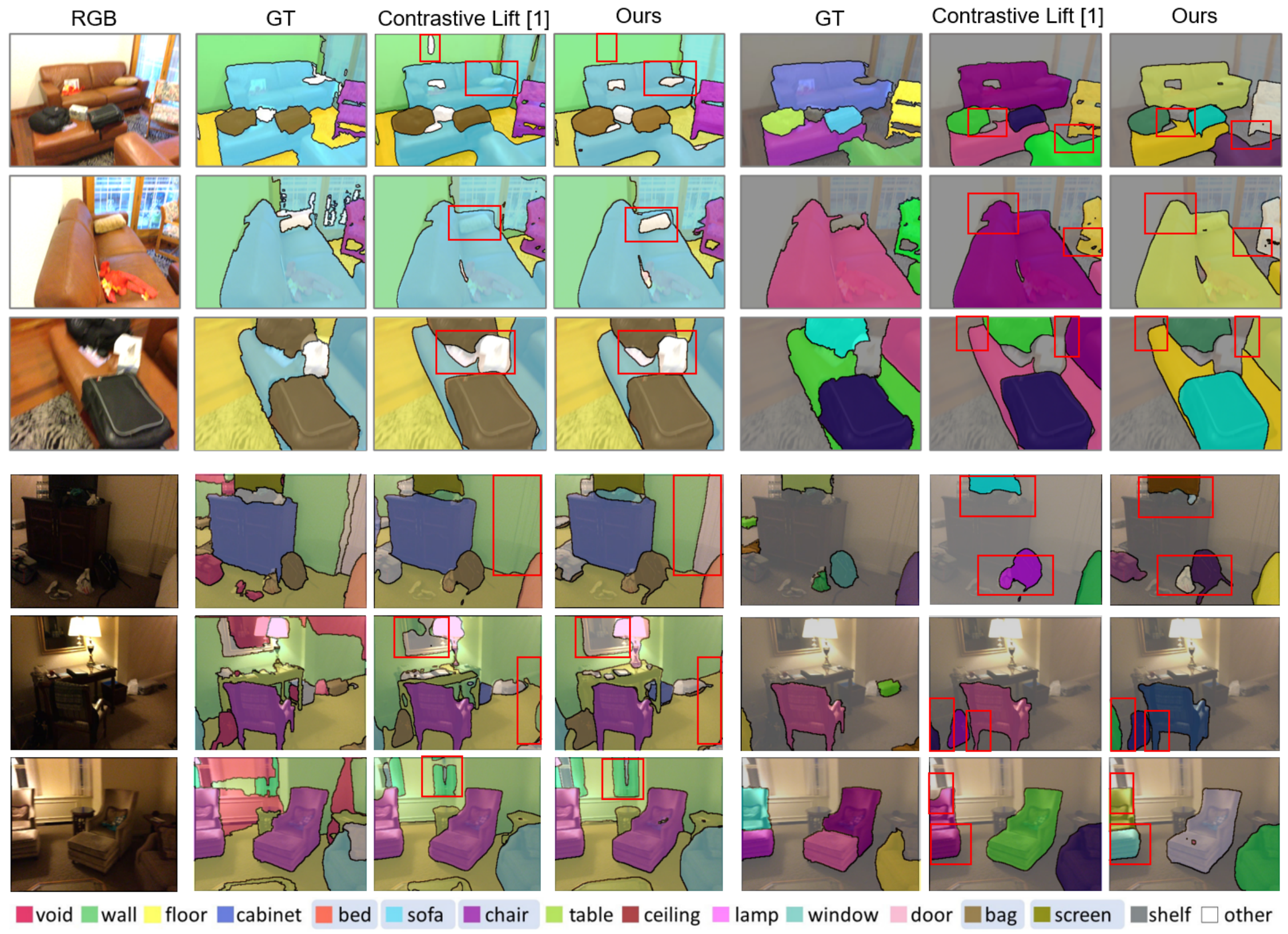}
  \caption{\textbf{Multi-View Comparisons against Contrastive Lift \cite{bhalgat2023contrastive}. } Columns 2-4: semantic segmentation. Columns 5-7: instance segmentation. Please zoom in to see the details.
  }
  \label{fig: multi view}
\end{figure}

In \cref{fig: multi view} we provide the segmentation comparisons from multiple camera views against Contrastive Lift \cite{bhalgat2023contrastive}, which can demonstrate that our segmentation is consistent across multiple views. On the other hand, it can also be observed that the segmentation results of \mname are closer to the actual geometry of the objects (the areas marked by the red boxes), which is compatible with the objective of \mname.

%
%